\documentclass[10pt,twocolumn,letterpaper]{article}

\usepackage{cvpr}              

\definecolor{cvprblue}{rgb}{0.21,0.49,0.74}
\usepackage[pagebackref,breaklinks,colorlinks,allcolors=cvprblue]{hyperref}
\usepackage{multirow}
\usepackage{makecell}

\usepackage{hyperref} 
\hypersetup{
    colorlinks=true,            
    linkcolor=red,              
    citecolor=green,            
    filecolor=mycustompurple,   
    urlcolor=magenta            
}

\def\paperID{8596} 
\def\confName{CVPR}
\def\confYear{2025}

\title{LightFC-X: Lightweight Convolutional Tracker for RGB-X Tracking}

\author{Yunfeng Li\\
\and
Bo Wang\\
\and
Ye Li\\
}

\begin{document}
\maketitle
\begin{abstract}
Despite great progress in multimodal tracking, these trackers remain too heavy and expensive for resource-constrained devices. To alleviate this problem, we propose LightFC-X, a family of lightweight convolutional RGB-X trackers that explores a unified convolutional architecture for lightweight multimodal tracking. Our core idea is to achieve lightweight cross-modal modeling and joint refinement of the multimodal features and the spatiotemporal appearance features of the target. Specifically, we propose a novel efficient cross-attention module (ECAM) and a novel spatiotemporal template aggregation module (STAM). The ECAM achieves lightweight cross-modal interaction of template-search area integrated feature with only 0.08M parameters. The STAM enhances the model's utilization of temporal information through module fine-tuning paradigm. Comprehensive experiments show that our LightFC-X achieves state-of-the-art performance and the optimal balance between parameters, performance, and speed. For example, LightFC-T-ST outperforms CMD by 4.3\% and 5.7\% in SR and PR on the LasHeR benchmark, which it achieves 2.6x reduction in parameters and 2.7x speedup. It runs in real-time on the CPU at a speed of 22 fps.  The code is available at \url{https://github.com/LiYunfengLYF/LightFC-X}.
\end{abstract}    
\section{Introduction}
\label{sec:intro}

Multimodal tracking uses a multimodal template to predict the target state in subsequent multimodal frames. It significantly improves the performance of visual trackers in challenging scenarios and has a wide range of potential applications \cite{rgbt1,rgbd2,rgbe3,rgbs50}.

In recent years, with the introduction of Transformer, multimodal trackers have made significant progress. However, these trackers inevitably become heavy and expensive. Take RGB-T tracking as an example, some methods insert integration modules into ViT, such as TBSI~\cite{tbsi} and CSTNet~\cite{cstnet}, their parameters and FLOPs are 202.0M and 144.0M, 82.5G and 74.7G, respectively. In addition, some methods like ViPT~\cite{vipt} and MPLT~\cite{mplt}, they use the prompt learning paradigm with only a few additional parameters, but their baseline OSTrack~\cite{ostrack} still has 92.1M parameters and 43.0G FLOPs. Similar problems exist for RGB-D and RGB-E tracking. The parameters and FLOPs of these trackers limit their application on resource-constrained devices. 

Although some MDNet trackers \cite{manet,dafnet,macnet} achieve lightweight multimodal tracking, their speed and performance is insufficient. In addition, CMD~\cite{cmd} explores how to design a lightweight DCF tracker through knowledge distillation in RGB-T tracking. However, there still exists a performance gap between CMD~\cite{cmd} and high-performance trackers. Furthermore, in RGB-D and RGB-E tracking, lightweight multimodal trackers have not received sufficient attention, resulting in challenges for the perception modules of many small intelligent vehicles (such as UAVs, etc.) to benefit from multimodal tracking.

\begin{figure}
\centering
\includegraphics[width=8.4cm]{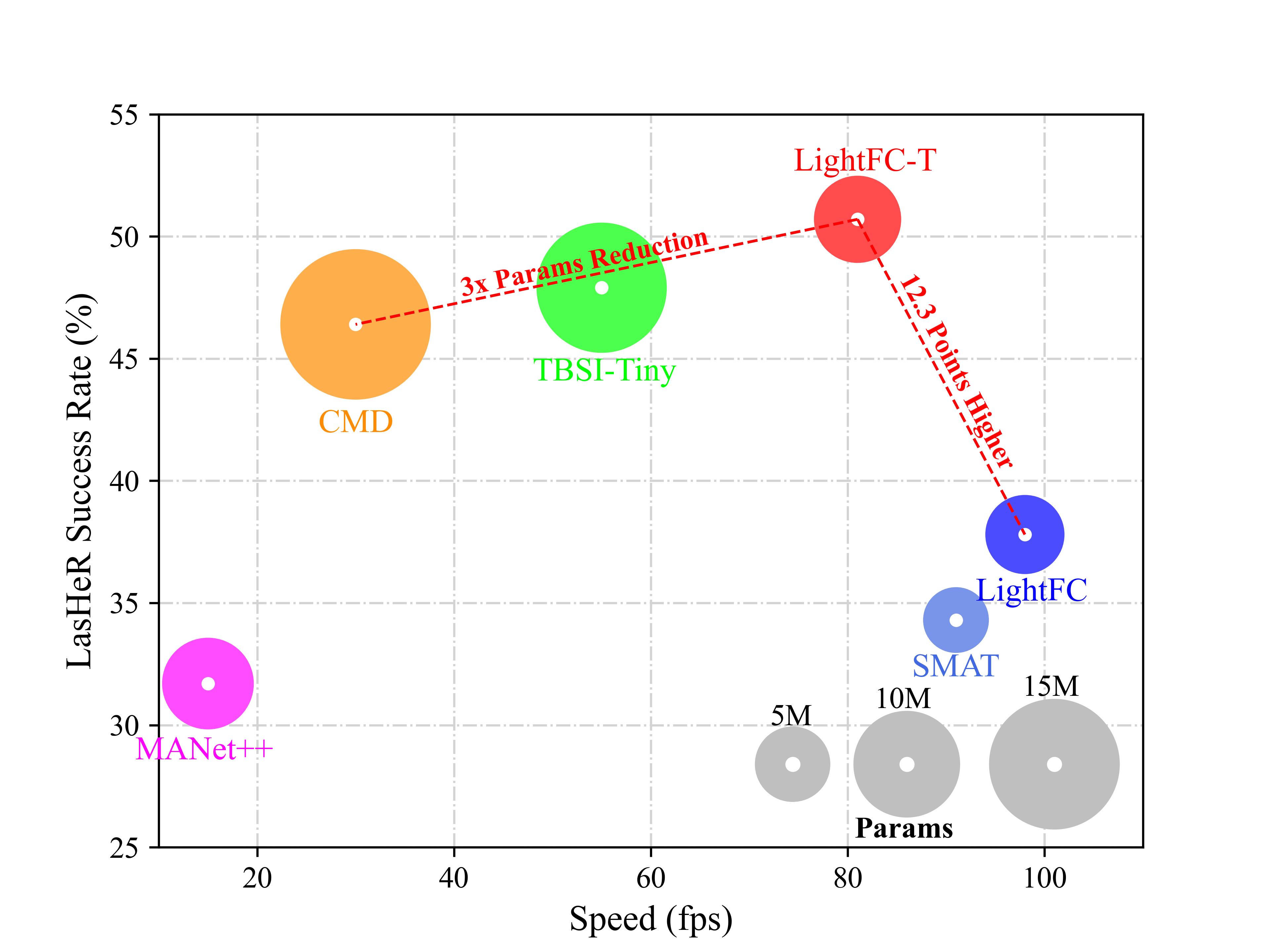}
\caption{Comparisons with state-of-the-art lightweight RGB-T trackers in terms of success rate, speed and parameters on LasHeR~\cite{lasher} benchmark. The proposed LightFC-T is superior than TBSI-Tiny~\cite{tbsi}, CMD~\cite{cmd}, and MANet++~\cite{manetpp}, while using much fewer parameters. }
\label{fig:bubblemap}
\end{figure}

To alleviate this problem, we aim to explore an efficient RGB-X tracker family using a unified tracking framework. A unified model will also facilitate the development of specific computing devices for acceleration. Therefore, the key issues are how to select an appropriate framework and how to achieve efficient cross-modal feature modeling.


For the first point, we first analyze the lightweight visual trackers of the last two years. With similar performance, LiteTrack-B4~\cite{litetrack} has 26.19M parameters and 6.78G FLOPs, which is 4.7x and 2.7x more than those of LightFC-vit~\cite{lightfc}, which has 5.50M parameters and 2.48G FLOPs. In addition, the operator optimization of convolution is sufficient, and most edge devices do not limit the inference efficiency of lightweight models. Although Transformer trackers display high speed on Nvidia edge devices, their speed may reduce on some embedded devices that lack attention operator optimization but are cheaper. Therefore, we adopt the convolutional framework of LightFC~\cite{lightfc} as our baseline.

For the second point, considering that the dimension of the lightweight features is small, the insertion of cross-modal feature interaction module does not significantly increase the model parameters. Moreover, some trackers \cite{tbsi,cstnet} show the benefits of inserting feature integration modules for cross-modal feature interaction, Therefore, we adopt the module insertion paradigm for cross-modal feature interaction.

In this work, we propose a family of lightweight RGB-X (Thermal, Depth, Event, and Sonar) trackers called LightFC-X. It includes a novel efficient cross-attention module (ECAM) and a novel spatiotemporal template aggregation module (STAM). The ECAM module takes the original multimodal features as Q, K, and V matrices, computes lightweight cross-modal cross-attention, and then performs joint spatial and channel integration on the multimodal features. The STAM module uses the first template as a fixed template and extracts dynamic templates in tracking, performing similar processing on the two templates as described above. For training, we first train the model with the ECAM module while omitting the STAM module to achieve spatial feature discrimination. Then, we freeze the parameters of the model and finetune the STAM module to model the spatiotemporal representation of the target.

Comprehensive experiments show that the proposed LightFC-X achieves state-of-the-art performance and an optimal balance between parameters, performance, and speed among current lightweight trackers. For example, for RGB-T tracking, LightFC-T and LightFC-T-ST surpass CMD~\cite{cmd} by 4.8\% and 5.7\% in PR, 3.7\% and 4.3\% in SR on the LasHeR~\cite{lasher}, while using 3x and 2.6x fewer parameters, respectively, and running 2.7x faster. In the DepthTrack~\cite{depthtrack} and VisEvent~\cite{visevent} benchmarks, LightFC-D-ST and LightFC-E-ST achieve similar performance to OSTrack~\cite{ostrack} while using 12x fewer parameters. Moreover, LightFC-X achieves real-time performance on a CPU at a speed of 22 \textit{fps}.

The contributions of our work are as follows:

(1) We propose a novel efficient cross-attention module (ECAM) with only 0.08M parameters. It achieves efficient cross-modal feature interaction and multimodal feature refinement.

(2) We propose a novel spatiotemporal template aggregation module (STAM). It achieves a fast transition from a spatial model to a spatiotemporal model through module fine-tuning paradigm.

(3) We propose LightFC-X, a family of lightweight RGB-X trackers. Comprehensive experiments demonstrate that LightFC-X achieves state-of-the-art performance on several multimodal tracking benchmarks.
\section{Related Work}
\label{sec:related}

\subsection{Lightweight Visual Object Tracking}
LightTrack~\cite{litetrack} uses Neural Architecture Search (NAS) technology to design a lightweight convolutional tracker. E.T.Track~\cite{ettrack} proposes an exemplar Transformer module for efficient tracking. FEAR~\cite{fear} proposes a hand-designed lightweight convolutional model and introduces a dual template update mechanism. MixFormerV2-S~\cite{mixformerv2s} proposes a one-stream Transformer framework based on knowledge distillation. HiT~\cite{hit} integrates the high-level semantics of the target features into the low-level features level by level in the Transformer model for achieving better features. SMAT~\cite{smat} proposes a separable mixed attention Transformer and an efficient self-attention module for lightweight tracking. LightFC~\cite{lightfc} explores the benefits of feature refinement for template-search area interacted feature and proposes a reparameterized prediction head. LiteTrack~\cite{litetrack} proposes a Transformer-based asynchronous feature extraction and interaction model. Although these methods are efficient, most of them are Transformer-based trackers, which inevitably increases their parameters. In this work, to achieve an optimal balance between parameters, FLOPs, performance, and speed, we adopt a convolutional architecture and we focus on making it more efficient.

\begin{figure*}
	\centering
    \includegraphics[width=17cm]{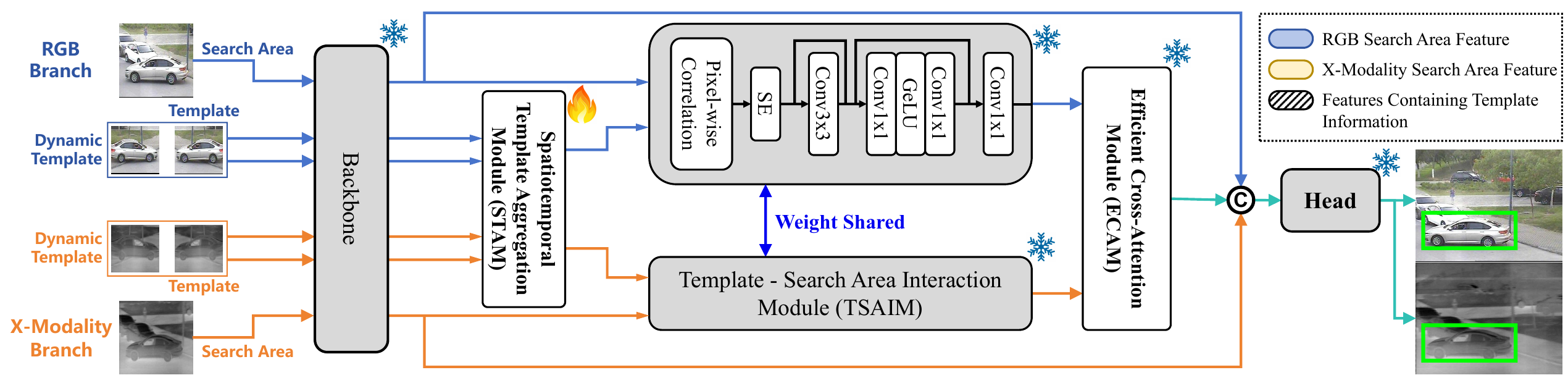}
	\caption{The overall framework of LightFC-X. It first obtains a spatiotemporal template feature of the target by aggregating the template and the dynamic template features. It then uses the TSAIM module to obtain two single-modality features. Our proposed ECAM module achieves lightweight interaction of template-search area integrated features. The cross-modal integrated feature is concatenated with the search area features from the two modalities and subsequently fed into the prediction head. During training, we first train the model that includes the ECAM module but excludes the STAM module. Afterward, we freeze the model and finetune the STAM module.}
\label{fig: framework}
\end{figure*}

\subsection{Lightweight Multimodal Tracking}
Some MDNet trackers for RGB-T tracking have fewer parameters, such as MANet~\cite{manet} has 7.28M parameters, DAFNet~\cite{dafnet} has 5.50M parameters, and MANet++~\cite{manetpp} has 7.38M parameters. However, shallow feature extraction networks limit the model's use of deep discriminative features. CMD~\cite{cmd} proposes a feature distillation module to facilitate a shallow network to learn multimodal feature representations of targets. TBSI~\cite{tbsi} provides a lightweight version, TBSI-Tiny, which inserts template search area bridging modules into ViT-Tiny. In addition, USTrack~\cite{ustrack} and FTOS~\cite{ftos} achieve high-speed GPU operation through knowledge distillation.
For other multimodal tracking, very few lightweight trackers are currently publicly reported. 
In this work, compared with these efficient multimodal trackers, LightFC-X not only achieves state-of-the-art performance but also achieves the optimal balance between parameters, performance, FLOPs, and speed.

\subsection{Spatiotemporal Template update}

UpdateNet~\cite{updatenet} proposes a convolutional template update network to achieve optimal template estimation for SiamFC~\cite{siamfc}. LTMU~\cite{ltmu} proposes a meta-updater to determine the conditions for template update. TCTrack~\cite{tctrack} proposes temporally adaptive convolution and Transformer modules. Stark-ST~\cite{stark} proposes a dynamic template usage and updating method within the Transformer framework. ODTrack~\cite{odtrack} models spatiotemporal information using continuously propagating trajectory tokens. AQATrack~\cite{aqatrack} uses an auto-regressive query to learn the spatiotemporal features of the target across  successive frames. TATrack~\cite{tatrack} proposes a spatiotemporal two-stream structure and online template-update method. However, these methods are developed for large trackers and are difficult to apply to lightweight trackers. FEAR~\cite{fear} proposes a template update method for lightweight trackers. However, this method utilizes linear interpolation for naive template updates. In this work, we use cross-attention to model the spatial feature associations between fixed and dynamic templates to achieve deep exploitation of temporal appearance features.

\section{Method}
\label{sec:method}

The overall framework is shown in Fig.~\ref{fig: framework}. First, we discuss the framework of LightFC-X (T for thermal, D for depth, E for event, and S for sonar). Then, we present the proposed ECAM module  and the STAM module. Finally, we introduce the prediction head and the training loss.

\subsection{Lightweight MultiModal Tracking framework}
\label{method rgbs}

We select LightFC \cite{lightfc} as our baseline to better balance parameters, performance, FLOPs, and speed. The modal of LightFC-S is slightly different due to the spatial misalignment of RGB-Sonar tracking, which we will discuss it separately. The inputs of our method are the RGB and X modalities (Thermal, Depth, Event, and Sonar) template images represented as  $z_{r}, z_{x} \in R^{H_{z}\times W_{z}\times 3}$ and the search area images are represented as $x_{r}, x_{x} \in R^{H_{x}\times W_{x}\times 3}$. We first use the backbone~\cite{tinyvit} to extract features and represent them as $Z_{r}, Z_{x} \in R^{C \times H_{z}^{f}\times W_{z}^{f}}$ and $X_{r}, X_{x} \in R^{C \times H_{x}^{f}\times W_{x}^{f}}$, where $C=160$ is the channels. $H_{z}^{f}=H_{z}/s, H_{x}^{f}=H_{x}/s$, where $s=16$ is the stride. During tracking, we use the proposed STAM module to aggregate the spatiotemporal feature of the target. The process is represented as $Z_{rt}=\phi(Z_{r}^{1}, Z_{r}^{i})$. Similarly, we obtain $Z_{xt}=\phi(Z_{x}^{1}, Z_{x}^{i})$.

First, we model the relationship between the template and the search area in different modalities, represented as:
\begin{equation}
A_{r}=Z_{rt}^{T}X_{r}, A_{x}=Z_{xt}^{T}X_{x}
\end{equation}
Then, we follow the framework of LightFC~\cite{lightfc} to refine the features. The refined features are represented as $X_{r}^{ts}, X_{x}^{ts} \in R^{C_{a} \times H_{x}^{f}\times W_{x}^{f}}$, where $C_{a}=96$.

The proposed efficient cross-attention module (ECAM) is represented as $X_{rx}^{st}=f(X_{r}^{ts}, X_{x}^{ts})$, where $X_{rx}^{st}$ denotes the cross-modal integrated feature. Then, we concatenate $X_{rx}^{st}$, $X_{r}$, and $X_{x}$ to supplement the original semantic lost during the fusion, represented as:
\begin{equation}
X_{rx}^{fusion}=[X_{rx}^{st}; X_{r}; X_{x}] \in R^{(2C+C_{a})\times H_{x}^{f}\times W_{x}^{f}}
\end{equation}
where $X_{fusion}$ is fed into the prediction head.

In RGB-Sonar (RGB-S) tracking, since the target is spatial misaligned in the RGB image and the sonar image~\cite{rgbs50}, we need to adjust the framework. The ECAM module for RGB-S tracking is denoted as $X_{r}^{xst}, X_{x}^{rst}=f^{'}(X_{r}^{ts}, X_{x}^{ts})$. Similarly, the feature concatenation is represented as:
\begin{equation}
\begin{split}
X_{i}^{fusion}=[X_{i}^{ist}; X_{i}] \in R^{C+C_{a}}, i\in(r,x)
\end{split}
\end{equation}
where $X_{i}^{fusion}, i\in(r,x)$ are input into the RGB prediction head and the sonar prediction head to predict the state of the target in two modalities, respectively.

\subsection{Efficient Cross-Attention Module}

\begin{figure}
\centering
\includegraphics[width=8.4cm]{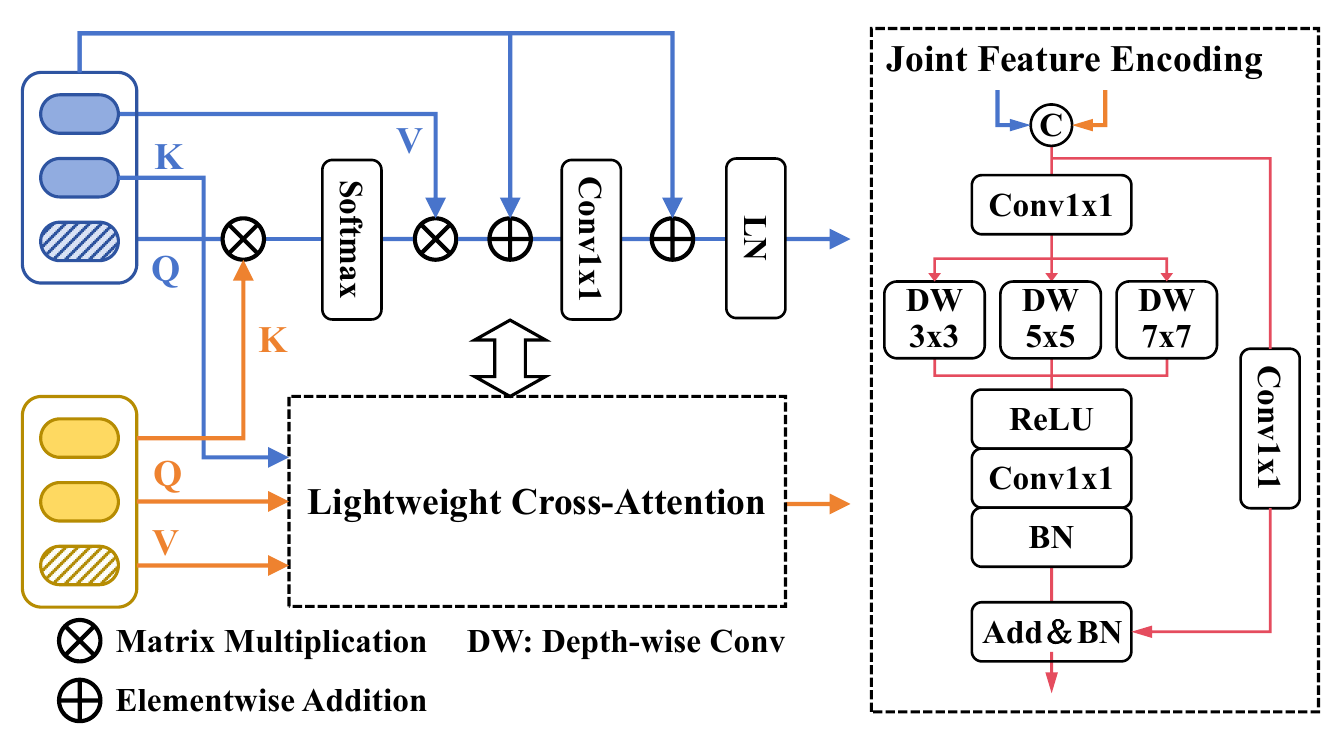}
\caption{Illustration of our proposed ECAM module. It contains a lightweight cross-attention layer and a joint feature encoding module.}
\label{fig:ecam}
\end{figure}

As shown in Fig.~\ref{fig:ecam}, our proposed efficient cross-attention module (ECAM) consists of a lightweight cross-attention layer and a joint feature encoding module. We insert the ECAM module between the template-search area feature interaction module and the prediction head.

\begin{figure}
\centering
\includegraphics[width=8.4cm]{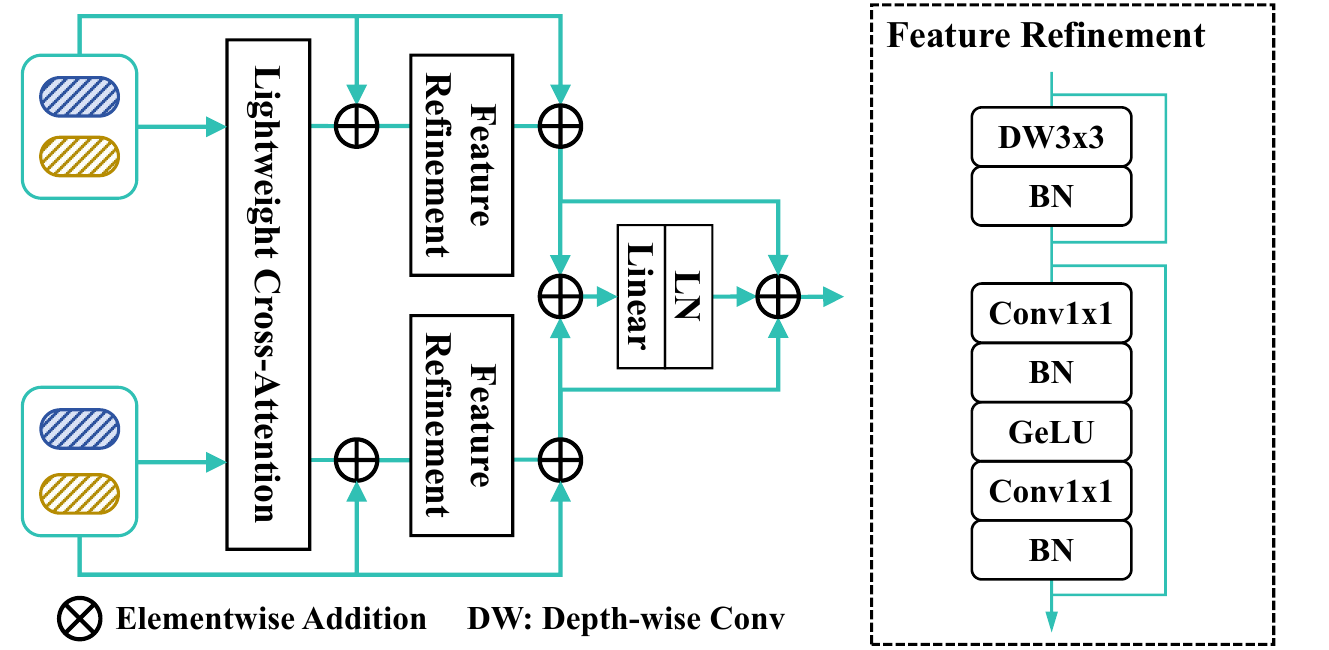}
\caption{Illustration of our proposed STAM module. It contains a lightweight cross-attention layer, two feature refinement modules, and a linear transformation layer.}
\label{fig: stam}
\end{figure}

\textbf{Lightweight Cross-Modal Integration.} 
We first convert the features $X_{r}^{ts}, X_{x}^{ts}$ into tokens $X_{r}', X_{x}'\in R^{N\times C_{a}}$, where $N=H_{x}^{f}W_{x}^{f}$. Then, we use an lightweight spatial cross-attention directly to the features, represented as:
\begin{equation}\small
\begin{split}
X_{r}^{crs} &=\text{Softmax}(\frac{Q_{r}K_{x}^{T}}{\sqrt{C}})V_{r}=\text{Softmax}(\frac{X_{r}'X_{x}'^{T}}{\sqrt{C}})X_{r}' \\
X_{x}^{crs} &=\text{Softmax}(\frac{Q_{x}K_{r}^{T}}{\sqrt{C}})V_{x}=\text{Softmax}(\frac{X_{x}'X_{r}'^{T}}{\sqrt{C}})X_{x}'
\end{split}
\end{equation}
where $Q, K, V$ are the Query, Key, and Value matrices. $Q_{x}K_{r}^{T}\in R^{1\times N\times N}$.

We use residual concatenation to complement the original semantic information of the target, represented as:
\begin{equation}
\begin{split}
X_{r}^{cross} = X_{r}'+X_{r}^{crs}, X_{x}^{cross} = X_{x}'+X_{x}^{crs}
\end{split}
\end{equation}
We then perform channel fusion on the features and use residual concatenation to supplement the original semantic, represented as:
\begin{equation}
\begin{split}
X_{i}^{eca} = \text{LN}^{\text{i}} (X_{i}'+\text{Conv}_{1\times 1}^{i}({X_{i}^{res}})) 
\end{split}
\end{equation}
where $i\in\{r,x\}$. Different superscripts indicate that they are independent layers.

\textbf{Joint Feature Encoding.} 
We use a joint feature encoding module to refine the features. First, we concatenate the multimodal features $X_{r}^{eca}$ and $X_{x}^{eca}$ to achieve joint encoding of the RGB modality and the X modality features, represented as $X_{rx}^{cat}=[X_{r}^{eca}; X_{x}^{eca}] \in R^{N\times 2C}$.

Then we downsample the features using a 1x1 convolutional layer, represented as:
\begin{equation}
X_{rx}^{down}=\text{Conv}_{1\times 1}^{down}(X_{rx}^{cat}) \in R^{N\times C}
\end{equation}
To improve the spatial representation of the target, we further perform a multi-level spatial aggregation as:
\begin{equation}
X_{rx}^{space}= \sum_{k\in (3,5,7)}\text{DWConv}_{k\times k}(X_{rx}^{space})
\end{equation}
To improve the nonlinearity of the model, we introduce the ReLU function to integrate the channel features as:
\begin{equation}
X_{rx}^{act}= \text{BN}(\text{Conv}_{1\times 1}(\text{ReLU}(X_{rx}^{space}))
\end{equation}
Where $\text{Conv}_{1\times 1}$ maps the feature from $R^{C}$ to $R^{C}$, $\text{BN}$ is the BatchNorm layer.

To complement the semantic information lost during encoding, we introduce downsampling and residual concatenation, represented as:
\begin{equation}
X_{rx}^{output}= \text{BN}(X_{rx}^{act}+\text{Conv}_{1\times 1}'(X_{rx}^{cat}))
\end{equation}
where $X_{rx}^{output}$ is the output feature of the proposed ECAM module. $\text{Conv}_{1\times 1}'$ is an unbiased convolutional layer.

For RGB-S tracking, we use two joint feature encoding modules to learn the refinement of the RGB feature and the sonar feature, respectively. Accordingly, the process is:
\begin{equation}
\begin{split}
X_{r}^{out}= \text{JFE}_{r}(X_{r}^{eca}), X_{x}^{out}=\text{JFE}_{x}(X_{x}^{eca})
\end{split}
\end{equation}
where $\text{JFE}$ represents the joint feature encoding module.

\subsection{Spatiotemporal Template Aggregation Module}
\label{section 3.3}
Our proposed spatiotemporal template aggregation module (STAM) includes a lightweight cross-attention layer, a feature integration module, and a linear transformation layer, as shown in Fig.~\ref{fig: stam}. 

\textbf{Spatiotemporal Feature Interaction.} The initial frame target template, considered as a fixed template is denoted as $Z_{r}^{1}$, and during tracking, the dynamic appearance template is denoted as $Z_{r}^{i}$.
\begin{equation}
    Z_{rt}^{1}, Z_{rt}^{i} = \text{CrossAttn}(Z_{r}^{1}, Z_{r}^{i})
\end{equation}
where $\text{CrossAttn}$ is the lightweight cross-attention above.

\textbf{Feature Refinement.} We then integrate the two template features using two decoupled spatial and channel enhancement layers, respectively. Since the feature processing is identical for both templates, we only show one of them. First, we model the local spatial relationships as:
\begin{equation}
    Z_{rt}^{is} = \text{DWConv}_{33}(Z_{rt}^{i})+Z_{rt}^{i}
\end{equation}
Then, we perform nonlinear channel integration as:
\begin{equation}
    Z_{rt}^{ic} = \text{Conv}_{11}^\text{down}(\text{GELU}(\text{Conv}_{11}^\text{up}(Z_{rt}^{is})))+Z_{rt}^{is}
\end{equation}
where $\text{Conv}_{11}^\text{up}$ maps the feature from $R^{C}$ to $R^{2C}$, $\text{GELU}$ represents the GeLU~\cite{gelu} function. Similarly, we get $Z_{rt}^{1c}$.

\textbf{Linear Transformation}. We then perform a joint linear transformation of $Z_{rt}^{is}$ and $Z_{rt}^{ic}$, denoted as:
\begin{equation}
    Z_{rt} = \text{LN}(\text{Linear}(Z_{rt}^{1c}+Z_{rt}^{ic})+Z_{rt}^{1c}+Z_{rt}^{ic})
\end{equation}
where $Z_{rt}$ is the output feature of the STAM module.

\textbf{Module Fine-Tuning.} During training, the joint training of the STAM module and tracker's model relies on the ability to discriminate based features from a dual-mode combination. However, due to the discrimination limitations of the lightweight model, this optimal discrimination during training may become suboptimal when tracking is disturbed. Therefore, we adopt a fine-tuning paradigm to freeze the tracker's model and train only the STAM module. This maintains the model's adaptability to individual template feature and enhances its ability to discriminate continuous changes in target appearance by incorporating additional temporal information.

\subsection{Head and Loss}
 
Following the design of LightFC~\cite{lightfc}, rep-center-head is adopted as our prediction head. For RGB-T, RGB-D, RGB-E tracking, we use only one prediction head. For RGB-S tracking, we use two prediction heads that have the same structure but do not share parameters. 

We use the weighted focal loss~\cite{weightfocalloss} and the GIoU loss~\cite{giou} to train the classification branch and the box regression branch, respectively. The total training loss is:
\begin{equation}
    L_{total}=L_{cls}+\lambda_{iou}L_{iou}+\lambda_{l_{1}}L_{1}
\end{equation}
where $\lambda_{iou}=2$ and $\lambda_{l_{1}}=5$.

\section{Experiments}
\label{sec:experiment}

\begin{table*}[t]
\centering
\footnotesize
\renewcommand{\arraystretch}{1.05}
\setlength{\tabcolsep}{0.8pt}
\begin{tabular}{c|ccccccccc|cccccc}
\hline
\multirow{2}{*}{} & \multicolumn{9}{c|}{non-lightweight}                                               & \multicolumn{6}{c}{lightweight}                                              \\ \cline{2-16} 
                  & \thead{CA3DMS \\~\cite{CA3DMS}} & \thead{SiamM\_Ds \\~\cite{vot19}} & \thead{Siam\_LTD \\~\cite{vot20}} & \thead{TSDM \\~\cite{tsdm}}  & \thead{DAL \\~\cite{DAL} }   & \thead{LTDSEd \\~\cite{vot19}} & \thead{ATCAIS \\~\cite{vot20}} & \thead{DDiMP \\~\cite{vot20}} & \thead{OSTrack \\~\cite{ostrack}} & \thead{HiT-Small \\~\cite{hit}} & \thead{LightFC \\~\cite{lightfc}} & \thead{LiteTrack \\~\cite{litetrack}} & \thead{LightFC-D \\ \ } & \thead{LightFC-D- \\N3} & \thead{LightFC-D- \\N3-ST} \\ \hline
F                 & 0.223  & 0.336     & 0.376     & 0.384 & 0.429 & 0.405  & 0.476  & 0.485 & \color{red}\textbf{0.529}   & 0.340     & 0.440   & 0.459     & 0.500     & 0.517 & \color{blue}\textbf{0.538} \\
Pr                & 0.218  & 0.264     & 0.342     & 0.393 & 0.512 & 0.382  & 0.455  & 0.469 & \color{red}\textbf{0.536}   & 0.333     & 0.450   & 0.452     & 0.537     & \color{blue}\textbf{0.558} & 0.557 \\
Re                & 0.228  & 0.463     & 0.418     & 0.376 & 0.369 & 0.430  & 0.500  & 0.503 & \color{red}\textbf{0.522}   & 0.348     & 0.430   & 0.467     & 0.468     & 0.481 & \color{blue}\textbf{0.520} \\ \hline
\end{tabular}

\caption{Overall performance on DepthTrack~\cite{depthtrack}. 
}
\label{table rgbd depthtrack}

\end{table*}
\begin{table*}[t]
\centering
\footnotesize
\renewcommand{\arraystretch}{1.05}
\setlength{\tabcolsep}{1.0pt}

\begin{tabular}{c|ccccccccc|cccccc}
\hline
\multirow{2}{*}{} & \multicolumn{9}{c|}{non-lightweight}                                                & \multicolumn{6}{c}{lightweight}                                              \\ \cline{2-16} 
                  & \thead{ATOM \\~\cite{atom}}  & \thead{DiMP \\~\cite{dimp}}  & \thead{ATCAIS \\~\cite{vot20}} & \thead{DRefine \\~\cite{vot21}} & \thead{KeepTrack\\~\cite{keeptrack}} & \thead{DMTrack \\~\cite{vot22}} & \thead{ProTrack \\~\cite{protrack}} & \thead{DeT \\~\cite{det}}   & \thead{OSTrack \\~\cite{ostrack}} & \thead{HiT-Small \\~\cite{hit}} & \thead{LightFC \\~\cite{lightfc}} & \thead{LiteTrack \\~\cite{litetrack}} & \thead{LightFC-D \\ \ } & \thead{LightFC-D- \\N3} & \thead{LightFC-D- \\N3-ST} \\ \hline
EAO               & 0.505 & 0.543 & 0.559  & 0.592   & 0.606     & 0.658   & 0.651    & 0.657 & \color{red}\textbf{0.676}   & 0.446     & 0.582   & 0.576     & 0.587     & 0.603        & \color{blue}\textbf{0.605}           \\
A                 & 0.698 & 0.703 & 0.761  & 0.775   & 0.753     & 0.758   & 0.801    & 0.760 & \color{red}\textbf{0.803}   & 0.705     & 0.776   & 0.768     & 0.757     & 0.760        & \color{blue}\textbf{0.760}           \\
R                 & 0.688 & 0.731 & 0.739  & 0.760   & 0.797     & 0.851   & 0.802    & 0.845 & \color{red}\textbf{0.833}   & 0.589     & 0.728   & 0.714     & 0.743     & 0.757        & \color{blue}\textbf{0.778}           \\ \hline
\end{tabular}
\caption{Overall performance on VOT-RGBD2022~\cite{vot22}. 
}
\label{table rgbd vot2022}
\end{table*}

\begin{table}[t]
\centering
\scriptsize
\renewcommand{\arraystretch}{1.05}
\setlength{\tabcolsep}{1.5pt}
\begin{tabular}{c|c|cc|cc|ccc|c|c}
\hline
\multirow{2}{*}{Methods} & \multirow{2}{*}{Year} & \multicolumn{2}{c|}{\thead{GTOT \\~\cite{gtot}}} & \multicolumn{2}{c|}{\thead{RGBT234 \\~\cite{rgbt234}}} & \multicolumn{3}{c|}{\thead{LasHeR \\~\cite{lasher}}} & \multirow{2}{*}{FPS} & \multirow{2}{*}{Params} \\ \cline{3-9}
                         &                       & MPR         & MSR         & MPR           & MSR          & PR      & NPR     & SR      &                      &                         \\ \hline
MANet~\cite{manet}      & 2019                  & 89.4        & 72.4        & 77.7          & 53.9         & 45.7    & 40.8    & 33.0    & 1                    & 7.28M                   \\
DAFNet~\cite{dafnet}    & 2019                  & 89.1        & 71.6        & 79.6          & 54.4         & 44.9    & 39.0    & 31.1    & 14                   & 5.50M                   \\
mfDiMP~\cite{mfdimp}   & 2019                  & 87.7        & 73.1        & 82.4          & 58.3         & 58.3    & 54.2    & 45.6    & 22                   & 175.82M                 \\
MACNet~\cite{macnet}    & 2020                  & 88.0        & 71.4        & 79.0          & 55.4         & 48.3    & 42.3    & 35.2    & 1                    & 14.86M                  \\
FANet~\cite{fanet}     & 2021                  & 90.1        & 72.1        & 79.4          & 53.9         & 48.2    & 42.5    & 34.3    & 12                   & 38.44M                  \\
SiamCDA~\cite{siamcda}  & 2021                  & 87.7        & 73.2        & 76.0          & 56.9         & -       & -       & -       & 24                   & 107.90M                 \\
MANet++~\cite{manetpp}  & 2021                  & 88.2        & 70.7        & 80.0          & 55.4         & 46.7    & 40.8    & 31.7    & 15                   & 7.38M                   \\
ADRNet~\cite{adrnet}    & 2021                  & 90.4        & \color{red}{\textbf{73.9}}  & 80.9          & 57.1         & -       & -       & -       & 15                   & 68.50M                  \\
MFGNet~\cite{mfgnet}    & 2022                  & 88.9        & 70.7        & 78.3          & 53.5         & -       & -       & -       & 3                    & 8.09M                   \\
APFNet~\cite{apfnet}    & 2022                  & \color{blue}{\textbf{90.5}} & \color{red}{\textbf{73.9}}        & 82.7          & 57.9         & 50.0    & -       & 36.2    & -                    & 15.01M                  \\
MFNet~\cite{mfnet}      & 2022                  & \color{red}{\textbf{90.7}}  & \color{blue}{\textbf{73.5}}        & \color{red}{\textbf{84.4} } & \color{blue}{\textbf{60.1}} & 59.7   & 55.4    & 46.7   & 18  & 81.01M                  \\\hline
CMD~\cite{cmd}          & 2023                  & 89.2        & 73.4        & 82.4          & 58.4         & 59.0    & 54.6    & 46.4    & 30                   & 19.90M                  \\ 
TBSI-Tiny~\cite{tbsi}   & 2023                  & -           & -           & -             & -            & 61.7    & 57.8    & 48.9    & 21                   & 14.90M                   \\ \hline
HiT-Tiny~\cite{hit}     & 2023                  & 49.8        & 44.1        & 47.6          & 35.6         & 29.8    & 26.1    & 26.7    & 170                  & 9.59M                   \\
HiT-Small~\cite{hit}    & 2023                  & 59.7        & 51.2        & 57.9          & 42.9         & 36.7    & 33.0    & 31.0    & 159                  & 11.03M                  \\
SMAT~\cite{smat}        & 2024                  & 70.9        & 58.7        & 64.8          & 48.2         & 41.7    & 38.1    & 34.3    & 91                   & 15.30M                   \\
LiteTrack~\cite{litetrack} & 2024               & 71.6        & 59.5        & 65.5          & 48.9         & 45.7    & 42.1    & 37.1    & 226                  & 26.18M                  \\
LightFC~\cite{lightfc}  & 2024                  & 67.7        & 54.2        & 71.9          & 51.8         & 48.6    & 43.8    & 37.8    & 98                   & 5.50M                   \\ \hline
LightFC-T                & -                     & 88.2        & 71.1        & \color{blue}{\textbf{83.2}}          & \color{red}{\textbf{60.3}}         & \color{blue}{\textbf{63.8}}   & \color{blue}{\textbf{59.9}}   & \color{blue}{\textbf{50.1} }  & 81                   & 6.68M                   \\ 
LightFC-T-ST             & -                     & 88.7        & 71.4        & \color{blue}{\textbf{83.4}}          & \color{red}{\textbf{60.3}}         & \color{red}{\textbf{64.7}}    & \color{red}{\textbf{60.8}}    & \color{red}{\textbf{50.7} }   & 81                   & 7.61M                   \\ \hline
\end{tabular}
\caption{Overall performance on GTOT~\cite{gtot}, RGBT234~\cite{rgbt234}, and LasHeR~\cite{lasher}. The best two results are shown in \color{red}red \color{black}and \color{blue}{blue} \color{black} fonts. Consistent with CMD~\cite{cmd}, we test the speed of the trackers on the GTX1080Ti GPU.}
\label{table rgbt}
\end{table}

\begin{table}[t]

\centering
\footnotesize
\renewcommand{\arraystretch}{1.05}
\setlength{\tabcolsep}{2.1pt}
\begin{tabular}{c|c|c|ccc|ccc}
\hline
\multirow{2}{*}{} & \multirow{2}{*}{Params} & \multirow{2}{*}{Type} & \multicolumn{3}{c|}{RGB} & \multicolumn{3}{c}{Sonar} \\ \cline{4-9} 
                  &                         &                       & SR     & PR     & NPR    & SR      & PR     & NPR    \\ \hline
SiamCAR~\cite{siamcar}       & 51.38M      & RGB                   & 22.0   & 19.0   & 24.5   & 36.9    & 64.0     & 45.6   \\
SiamBAN~\cite{siamban}       & 53.93M      & RGB                   & 30.8   & 31.5   & 33.9   & 41.6    & 65.8   & 51.6   \\
TransT~\cite{transt}         & 18.54M      & RGB                   & 28.0   & 29.0   & 31.9   & 42.9    & 65.1   & 50.5   \\
Stark-S50~\cite{stark}      & 28.10M      & RGB                   & 35.0   & 35.8   & 40.0   & 40.2    & 60.5   & 50.1   \\
Stark-ST50~\cite{stark}     & 28.23M      & RGB                   & 35.2   & 35.4   & 39.4   & 41.1    & 62.3   & 51.4   \\
TransT-SLT~\cite{transtslt}  & 18.54M      & RGB                   & 27.9   & 28.9   & 31.7   & 42.9    & 65.1   & 50.5   \\
OSTrack256~\cite{ostrack}    & 92.12M      & RGB                   & 69.3   & 73.5   & 75.9   & 40.9    & 54.6   & 51.6   \\
SCANet~\cite{rgbs50}         & 112.77M     & RGBS                  & \color{red}\textbf{71.2}   & \color{red}\textbf{75.6}   & \color{red}\textbf{78.0}   & \color{red}\textbf{50.3}    & \color{red}\textbf{66.2}   & \color{red}\textbf{62.1}   \\ \hline
HiT-Tiny~\cite{hit}          & 9.59M                & RGB                   & 27.6   & 25.6   & 35.3   & 25.7    & 36.1   & 39.9   \\
HiT-Small~\cite{hit}         & 11.03M               & RGB                   & 30.8   & 28.2   & 37.2   & 36.7    & 54.7   & 48.6   \\
LightFC~\cite{lightfc}       & 5.50M             & RGB                   & 56.2   & 58.7   & 61.4   & 22.2    & 28.1   & 26.4   \\ \hline
LightFC-S                    & 10.40M                 & RGBS                  & \color{blue}\textbf{58.6}   & \color{blue}\textbf{59.6}   & \color{blue}\textbf{65.0}   & \color{blue}\textbf{39.5}    & \color{blue}\textbf{58.3}   & \color{blue}\textbf{52.0}   \\ \hline
\end{tabular}
\caption{Overall performance on RGBS50~\cite{rgbs50}. 
}
\label{table rgbs}
\end{table}

\begin{table*}[t]
\centering
\footnotesize
\renewcommand{\arraystretch}{1.05}
\setlength{\tabcolsep}{0.9pt}
\begin{tabular}{c|ccccccccc|cccccc}
\hline
\multirow{2}{*}{} & \multicolumn{9}{c|}{non-lightweight}                                                & \multicolumn{6}{c}{lightweight}                                       \\ \cline{2-16} 
                  & \thead{SiamCAR \\~\cite{siamcar}} & \thead{MDNet \\~\cite{mdnet}} & \thead{StarkS50 \\~\cite{stark}} & \thead{PrDiMP \\~\cite{prdimp}} & \thead{LTMU \\~\cite{ltmu}} & \thead{ProTrack \\~\cite{protrack}} & \thead{TransT\\~\cite{transt}} & \thead{SiamRCNN \\~\cite{siamrcnn}} & \thead{OSTrack \\~\cite{ostrack}} & \thead{HiT-Tiny \\~\cite{hit}} & \thead{HiT-Small \\~\cite{hit}} & \thead{LightFC \\~\cite{lightfc}} & \thead{LiteTrack \\~\cite{litetrack}} & \thead{LightFC-E \\ \ } & \thead{LightFC-E- \\ ST } \\ \hline
SR                & 42.0    & 42.6  & 44.6     & 45.3   & 45.9 & 47.1     & 47.4   & 49.9     & \color{red}{\textbf{53.4}}    & 35.8     & 40.0      & 47.2    & 43.1      & 52.9      & \color{blue}{\textbf{53.2}}         \\
PR                & 59.9    & 66.1  & 61.2     & 64.4   & 65.5 & 63.2     & 65.0   & 65.9     & \color{red}{\textbf{69.5}}    & 45.8     & 51.9      & 62.9    & 55.4      & 68.7      & \color{blue}{\textbf{69.1}}         \\ \hline
\end{tabular}
\caption{Overall performance on VisEvent test set~\cite{visevent}.}
\label{table rgbe}
\end{table*}

\begin{table}[]
\centering
\footnotesize
\renewcommand{\arraystretch}{1.05}
\setlength{\tabcolsep}{1.5pt}

\begin{tabular}{l|c|ccc|cccc}
\hline
\multirow{2}{*}{} & \multirow{2}{*}{Params} & \multicolumn{3}{c|}{PyTorch} & \multicolumn{4}{c}{Onnx }   \\ \cline{3-9} 
                  &                         & CPU     & GPU1     & GPU2    & CPU & GPU1 & GPU2 & Xavier \\ \hline
LightFC-X         & 6.68M                   & 22      & 34       & 169     & 28  & 45   & 178  & 51     \\
LightFC-D-N3      & 6.91M                   & 18      & 27       & 156     & 25  & 37   & 165  & 46     \\
LightFC-S         & 10.40M                  & 17      & 24       & 150     & 22  & 34   & 159  & 42     \\ \hline
\end{tabular}
\caption{Comparison of parameters and speed. The CPU is Intel I9-12900KF, the GPU1 is GTX1050Ti, the GPU2 is GTX3090Ti, and the Xavier is NVIDIA Jetson AGX Xavier.}
\label{table params and flops and speed}
\end{table}

\begin{table}[t]
\centering
\footnotesize
\renewcommand{\arraystretch}{1.05}
\setlength{\tabcolsep}{2.5pt}

\begin{tabular}{clll|ccc|cc}
\hline
\multicolumn{4}{c|}{\multirow{2}{*}{}}                                                    & \multicolumn{3}{c|}{LasHeR~\cite{lasher}} & \multicolumn{2}{c}{VisEvent~\cite{visevent}} \\ \cline{5-9} 
\multicolumn{4}{c|}{}                                                                     & PR      & NPR     & SR      & PR            & SR           \\ \hline
\multicolumn{4}{c|}{Baseline}                                                             & 48.6    & 43.8    & 37.8    & 62.9          & 47.2         \\ \hline
\multicolumn{2}{c|}{\multirow{5}{*}{ECAM}} & \multicolumn{2}{l|}{+ Cross Attention}       & 38.4    & 36.8    & 29.7    & 41.4          & 30.3         \\
\multicolumn{2}{l|}{}                      & \multicolumn{2}{l|}{+ Conv1x1 Integration}   & 55.3    & 52.1    & 43.5    & 55.9          & 41.2         \\
\multicolumn{2}{l|}{}                      & \multicolumn{2}{l|}{+ Spatial Aggregation}   & 58.3    & 54.9    & 46.0    & 62.1          & 44.6         \\
\multicolumn{2}{l|}{}                      & \multicolumn{2}{l|}{+ Channel Aggregation}   & 58.9    & 55.3    & 46.5    & 64.4          & 47.1         \\
\multicolumn{2}{l|}{}                      & \multicolumn{2}{l|}{+ Residual Connection}   & 58.6    & 55.4    & 46.0    & 66.0          & 48.1         \\ \hline
\multicolumn{2}{c|}{Framework}             & \multicolumn{2}{l|}{+ Feature Concatenation} & \textbf{63.8} & \textbf{59.9} & \textbf{50.1} & \textbf{68.7} & \textbf{52.9}         \\ \hline
\multicolumn{2}{c|}{\multirow{3}{*}{STAM}} & \multicolumn{2}{l|}{+ Cross-Attention}       & 63.9    & 60.3    & 50.4    & 68.3          & 52.1         \\
\multicolumn{2}{l|}{}                      & \multicolumn{2}{l|}{+ Feature Integration}   & 64.4    & 60.5    & 50.5    & 68.6          & 52.9         \\
\multicolumn{2}{l|}{}                      & \multicolumn{2}{l|}{+ Linear Layer}          & \textbf{64.7} & \textbf{60.8} & \textbf{50.7} & \textbf{69.1} & \textbf{53.2}         \\ \hline
\end{tabular}
\caption{Ablation of our proposed LightFC-X.}
\label{table ablation ecm}
\end{table}
\begin{table}[]
\centering
\footnotesize
\renewcommand{\arraystretch}{1.05}
\setlength{\tabcolsep}{1.1pt}

\begin{tabular}{c|c|c|ccc|ccc|cc}
\hline
\multirow{2}{*}{} & \multirow{2}{*}{Params} & \multirow{2}{*}{FPS} & \multicolumn{3}{c|}{DepthTrack~\cite{depthtrack}} & \multicolumn{3}{c|}{LasHeR~\cite{lasher}} & \multicolumn{2}{c}{VisEvent~\cite{visevent}} \\ \cline{4-11} 
                  &                         &                      & F            & Pr           & Re      & PR      & NPR     & SR      & PR            & SR           \\ \hline
+0x ECAM          & 6.60M                   & 178                  & 47.9         & 51.9         & 44.5    & 60.0    & 56.1    & 46.8    & 67.5          & 51.1         \\
+1x ECAM          & 6.68M                   & 169                  & \textbf{49.3}& \textbf{57.2}& \textbf{43.3} & \textbf{63.8} & \textbf{59.9} & \textbf{50.1} & \textbf{69.1} & \textbf{53.2}         \\
+2x ECAM          & 6.75M                   & 163                  & 47.6         & 53.5         & 42.8    & 60.2    & 56.4    & 47.2    & 67.7          & 52.2         \\
+3x ECAM          & 6.83M                   & 156                  & \textbf{51.7}& \textbf{55.8}& \textbf{48.1}    & 60.2    & 56.7    & 47.5    & 68.1          & 52.5         \\ 
+4x ECAM          & 6.91M                   & 153                  & 51.3         & 55.6         & 49.3    & 58.2 & 54.9  & 46.4    & 67.5          & 51.2        \\ \hline
\end{tabular}
\caption{Ablation of different number of the ECAM module. }
\label{table ablation number ecm}
\end{table}

\subsection{Implementation Details}
The training set includes DepthTrack~\cite{depthtrack} for RGB-D tracking, LasHeR~\cite{lasher} for RGB-T tracking, and VisEvent~\cite{visevent} for RGB-E tracking. In addition, we follow~\cite{rgbs50} to train our LightFC-S using the LaSOT~\cite{lasot}, GOT10K~\cite{got10k}, TrackingNet~\cite{trackingnet}, COCO~\cite{coco}, and SarDet~\cite{sardet} datasets. The training platform consists of two NVIDIA RTX A6000 GPUs. The training consists of two phases. In the first phase we train the LightFC-X model without the STAM module. The training takes 45 epochs with a global batch size of 64. Each epoch contains 60k pairs of samples. The AdamW~\cite{adamw} optimizer is used with a weight decay of $10^{-4}$. The learning rate is set to $2\times 10^{-4}$ and decreased by 10 times after 10 epochs. In the second phase, we freeze all parameters of LightFC-X and train only the STAM module. The number of epochs is 15. All other training parameters are unchanged. The template update interval and the update confidence threshold of the tracking pipeline are uniformly set to 400 and 0.7 for multimodal tracking benchmarks.

\subsection{Evaluation Datasets and Metrics}
Our test set includes LasHeR~\cite{lasher}, RGBT234~\cite{rgbt234}, and GTOT~\cite{gtot} for RGB-Thermal tracking; DepthTrack~\cite{depthtrack} and VOT-RGBD2022~\cite{vot22} for RGB-Depth tracking; VisEvent~\cite{visevent} for RGB-Event tracking;  RGBS50~\cite{rgbs50} for RGB-Sonar tracking.

We evaluate our trackers using maximum precision rate (MPR) and maximum success rate (MSR) for GTOT~\cite{gtot} and RGBT234~\cite{rgbt234}, precision rate (PR) and success rate (SR) for LasHeR~\cite{lasher} and VisEvent~\cite{visevent}, F1 Score (F-Score), precision (Pr), and recall (Re) for DepthTrack~\cite{depthtrack}, expected average overlap (EAO), accuracy (A), and recall (R) for VOT-RGBD2022~\cite{vot22}, and RGBS50~\cite{rgbs50} with SR, PR, and normalized precision rate (NPR).

\subsection{Comparison with the state-of-the-art}

\

\textbf{DepthTrack.} DepthTrack~\cite{depthtrack} is a large-scale RGB-D tracking benchmark with 50 test sequences. 
As shown in Tab.~\ref{table rgbd depthtrack}, LightFC-D surpasses DDiMP~\cite{vot20} by 1.5\% and 6.8\% in F-Score and Pr, respectively. 
The F-score and Pr of LightFC-D-N3-ST are 53.8\% and 55.7\%, being 0.9\% and 2.1\% higher than OSTrack~\cite{ostrack}, while using 11.9x fewer parameters (7.76 \textit{v.s.} 92.18 M). 
Moreover, it achieves a gain of 9.8\% and 10.7\% in F-Score and Pr over the baseline LightFC~\cite{lightfc}. Overall, LightFC-D, LightFC-D-N3, and LightFC-D-N3-ST achieve state-of-the-art performance in lightweight RGB-D tracking. 

\textbf{VOT-RGBD2022.} VOT-RGBD2022~\cite{vot22} is a challenging short-term benchmark containing 127 test sequences. The results are shown in Tab.~\ref{table rgbd vot2022}. LightFC-D-N3 achieves similar performance to KeepTrack~\cite{keeptrack}. In addition, it outperforms LiteTrack~\cite{litetrack} by 2.9\% and 6.4\% in the EAO and R, respectively. In general, LightFC-D-N3-ST achieves competitive performance.

\textbf{LasHeR.} LasHeR~\cite{lasher} is a large-scale RGB-T tracking benchmark that comprises 245 test sequences. As shown in Tab.~\ref{table rgbt}. LightFC-T achieves superior performance compared to existing efficient RGB-T trackers. For example, LightFC-T outperforms CMD~\cite{cmd} by 3.7\% and 4.8\% in SR and PR, respectively, while using nearly 3x fewer parameters (6.68 \textit{v.s.} 19.90 M) and running 2.7x faster (81 \textit{v.s.} 30 \textit{fps}) respectively. Compared to MANet++ \cite{manetpp}, which has a similar number of parameters, LightFC-T surpasses it by 18.4\% in SR and 17.1\% in PR, while running 5.4x faster (81 \textit{v.s. }15 \textit{fps}). Compared to LightFC~\cite{lightfc}, LightFC-T outperforms it by 12.3\% and 15.2\% in SR and PR, respectively. Overall, LightFC-T and LightFC-T-ST achieve state-of-the-art performance among lightweight trackers for RGB-T tracking. 

\textbf{RGBT234.} RGBT234~\cite{rgbt234} is a large RGB-T tracking benchmark that contains 234 test sequences. As shown in Tab.~\ref{table rgbt}, the MPR and MSR of LightFC-T-ST are 1.0\% and 1.9\% higher than those of CMD~\cite{cmd}, respectively. Additionally, the MSR of LightFC-T-ST is comparable to MFNet~\cite{mfnet}, while using 10.6x fewer parameters (7.61 \textit{v.s.} 81.01 M) and running 4.5x faster (81 \textit{v.s. }18 \textit{fps}). In general, LightFC-T and LightFC-T-ST outperform other lightweight trackers in RGB-T tracking.

\textbf{GTOT.} GTOT~\cite{gtot} is a classic short-term RGB-T tracking benchmark comprising 50 sequences. Tab.~\ref{table rgbt} presents that LightFC-T achieves better MPR (88.2\%) and MSR (71.1\%), being  16.6\% and 11.6\% higher than those of LiteTrack-B4~\cite{litetrack}, while using nearly 4x fewer parameters (6.68 \textit{v.s.} 26.18 M). Overall, LightFC-T and LightFC-T-ST achieve competitive performance.

\textbf{VisEvent.} VisEvent~\cite{visevent} is a large-scale RGB-E tracking benchmark comprising 323 test sequences. As shown in Tab.~\ref{table rgbe}, LightFC-E achieves better SR (52.9\%) and PR (68.7\%), being 3.0\% and 2.8\% higher than those of SiamRCNN~\cite{siamrcnn}. LightFC-E-ST achieves performance close to that of OSTrack~\cite{ostrack}, while using 13.8x fewer parameters (6.68 \textit{v.s.} 92.18 M). Overall, LightFC-E and LightFC-E-ST demonstrate state-of-the-art performance for lightweight RGB-E tracking.

\textbf{RGBS50.} RGBS50~\cite{rgbs50} is a classic RGB-Sonar tracking benchmark comprising 50 sequences. As shown in Tab.~\ref{table rgbs}, for RGB sequences, LightFC-S outperforms LightFC~\cite{lightfc} by 2.4\% and 3.6\% in SR and NPR. For sonar sequences, compared to LightFC~\cite{lightfc}, LightFC-S improves the SR and PR by 17.3\% and 30.2\%, respectively. The results show the effectiveness of our method in spatially misaligned multimodal tracking. Overall, LightFC-S achieves state-of-the-art performance in lightweight RGB-S tracking.

\textbf{Params and FLOPs.} Tab.~\ref{table params and flops and speed} presents the speeds of the proposed trackers. On the resource-constrained GPU GTX1050Ti, LightFC-X, LightFC-D-N3, and LightFC-S run at 34, 27, and 24 \textit{fps}, respectively. Before deployment, LightFC-X runs in real-time on the CPU Intel I9-12900KF at 22 \textit{fps}. After deployment, LightFC-X, LightFC-D-N3, and LightFC-S all achieve real-time speed on the CPU. On the Xavier edge device, LightFC-X, LightFC-D-N3, and LightFC-S run at 51, 46, and 42 \textit{fps}, respectively. Overall, the LightFC-X family shows significant potential for practical deployment.

\subsection{Ablation Studies}

\ 

\textbf{Component Analysis.} We evaluate the contribution of each component of the proposed modules. As shown in Tab.~\ref{table ablation ecm}, each component of the ECAM and STAM modules contributes to the performance improvement. In particular, within the tracking framework, concatenating the original features from the search area with the output features of the ECAM module effectively complements the original semantic features of the target, thereby improving the performance of the prediction head.

\textbf{Number of ECAM Modules.} We evaluate the effect of inserting different numbers of ECAM modules on performance. As shown in Tab.~\ref{table ablation number ecm}, the ECAM module only has 0.08M parameters. LightFC-T and LightFC-E achieve the best performance when one ECAM module is inserted. For RGB-D tracking, performance is optimal when three ECAM modules are inserted. Excessive insertion of ECAM modules leads to a decline in model performance. Therefore, we present two versions for RGB-D tracking. LightFC-D shares the same structure as LightFC-T and LightFC-E. LightFC-D-N3 achieves superior performance with a slight increase in parameters.

\begin{figure}[t]
\centering
\includegraphics[width=7.8cm]{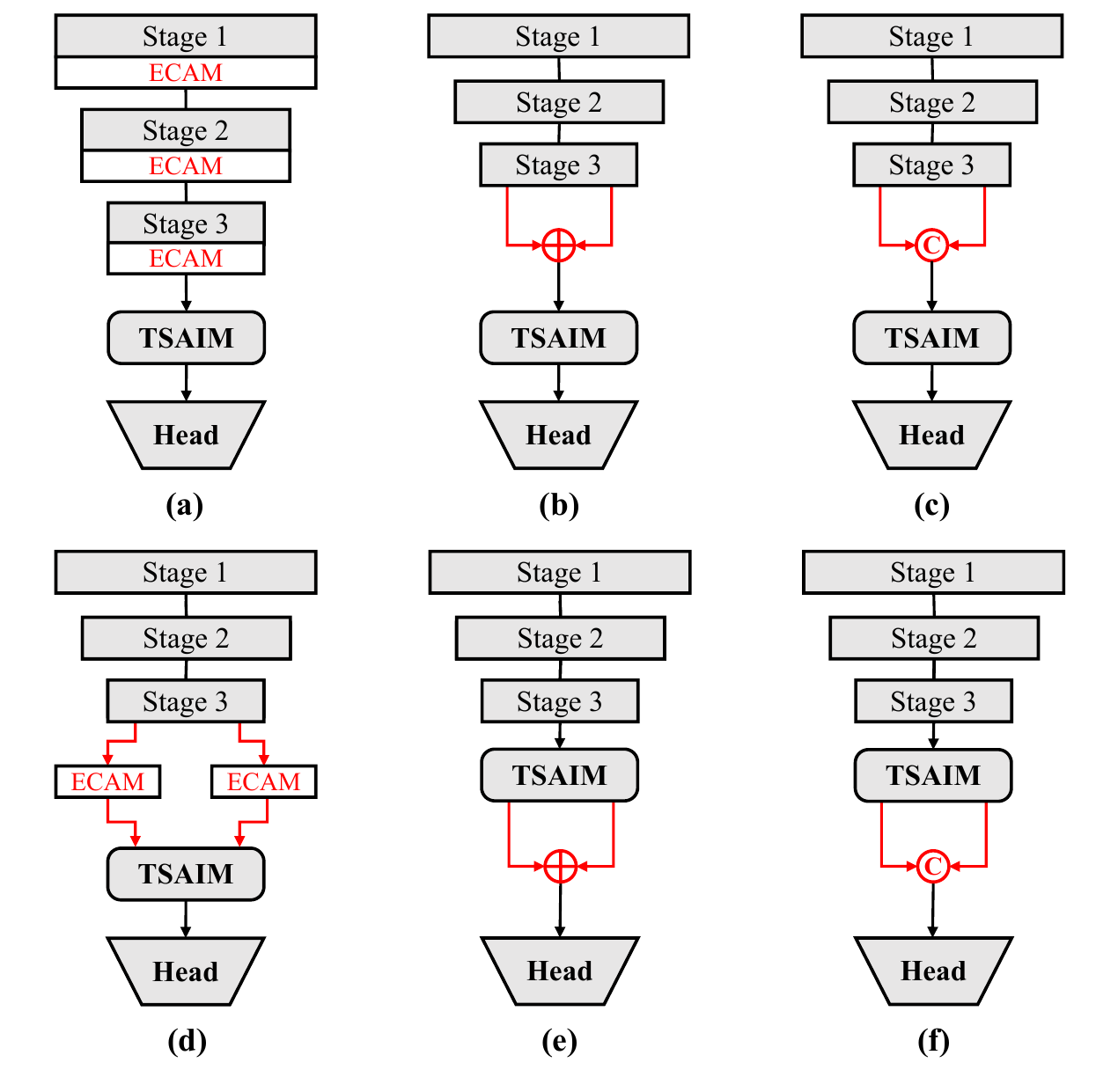}
\caption{Different variants of LightFC-X. "TSAIM" denotes the template-search area interaction module. (a) Integrating multimodal features in the Backbone. (b-d) Integrating multimodal features before TSAIM module. (e-f) Integrating multimodal features after TSAIM module.}
\label{fig:variant}
\end{figure}

\begin{table}[t]
\centering
\footnotesize
\renewcommand{\arraystretch}{1.05}
\setlength{\tabcolsep}{2.1pt}

\begin{tabular}{c|c|c|ccc|ccc|cc}
\hline
\multirow{2}{*}{} & \multirow{2}{*}{Params} & \multirow{2}{*}{FPS} & \multicolumn{3}{c|}{\thead{DepthTrack \\~\cite{depthtrack}}} & \multicolumn{3}{c|}{\thead{LasHeR \\~\cite{lasher}}} & \multicolumn{2}{c}{\thead{VisEvent \\~\cite{visevent}}} \\ \cline{4-11} 
                  &                         &                      & F         & Pr       & Re       & PR      & NPR     & SR      & PR            & SR           \\ \hline
LightFC           & 5.50M                   & 202                  & 44.0      & 45.0     & 43.0     & 48.6    & 43.8    & 37.8    & 62.9          & 47.2         \\ \hline
variant(a)        & 7.20M                   & 145                  & 43.3      & 45.1     & 41.7     & 56.6    & 53.4    & 45.4    & 66.8          & 51.3         \\
variant(b)        & 6.60M                   & 188                  & 48.4      & 53.1     & 48.4     & 55.8    & 52.5    & 44.1    & 65.7          & 49.9         \\
variant(c)        & 6.60M                   & 189                  & 48.8      & 55.1     & 43.8     & 59.8    & 56.3    & 47.4    & 65.8          & 50.0         \\
variant(d)        & 6.80M                   & 171                  & 48.2      & 51.9     & 48.2     & 55.7    & 52.1    & 44.2    & 63.0          & 48.3         \\
variant(e)        & 6.60M                   & 180                  & 47.9      & 51.9     & 44.5     & 57.3    & 53.8    & 45.3    & 67.7          & 51.4         \\
variant(f)        & 7.27M                   & 182                  & 49.3      & 57.2     & 43.3     & 57.7    & 54.2    & 45.3    & 67.9          & 51.3         \\ \hline
LightFC-X         & 6.68M                   & 169                  & \textbf{50.0} & \textbf{53.7} & \textbf{46.8} & \textbf{63.8} & \textbf{59.9} & \textbf{50.1} & \textbf{68.7} & \textbf{52.9} \\ \hline
\end{tabular}
\caption{Ablation of different variants of LightFC-X.}
\label{table ablation variant}
\end{table}

\begin{table}[t]
\centering
\footnotesize
\renewcommand{\arraystretch}{1.1}
\setlength{\tabcolsep}{3.1pt}

\begin{tabular}{c|c|ccc|ccc|cc}
\hline
\multirow{2}{*}{}     & \multirow{2}{*}{Freeze} & \multicolumn{3}{c|}{\thead{DepthTrack \\~\cite{depthtrack}}} & \multicolumn{3}{c|}{\thead{LasHeR \\~\cite{lasher}}} & \multicolumn{2}{c}{\thead{VisEvent \\~\cite{visevent}}} \\ \cline{3-10} 
                      &                         & F             & Pr            & Re               & PR      & NPR     & SR      & PR            & SR           \\ \hline
\multirow{2}{*}{STAM} &                         & 48.5          & 55.2          & 43.3             & 58.9    & 55.4    & 46.4    & 67.4          & 51.6       \\
                      & \checkmark              & \textbf{53.8} & \textbf{55.7} & \textbf{52.0}    & \textbf{64.7}    & \textbf{60.8}    & \textbf{50.7}    & \textbf{69.1}          & \textbf{53.2}         \\ \hline
\end{tabular}
\caption{Ablation of different STAM module training paradigms.}
\label{table ablation stam}
\end{table}

\textbf{Different Framework Variants.} We explore different framework variants of our LightFC-X. As shown in Fig.~\ref{fig:variant},  variant (a) aims to explore the cross-modal feature integration in the backbone. Variants (b-d) aim to explore cross-modal feature integration before the template-search area interaction. Variants (e-f) and LightFC-X aim to explore cross-modal feature integration after the template-search area interaction. The results are shown in Tab.~\ref{table ablation variant}, LightFC-X achieves state-of-the-art performance among all variants. This shows that cross-modal feature integration after template-search area interaction is more suitable for lightweight multimodal Siamese tracker.

\textbf{Training paradigm of STAM module.} We explore different training paradigms of our STAM module on LightFC-D-N3, LightFC-T, and LightFC-E. As shown in Tab.~\ref{table ablation stam}, freezing the tracking model during the training of the STAM module effectively improves the tracking performance.

\subsection{Visualization}
\begin{figure}
\centering
\includegraphics[width=7.4cm]{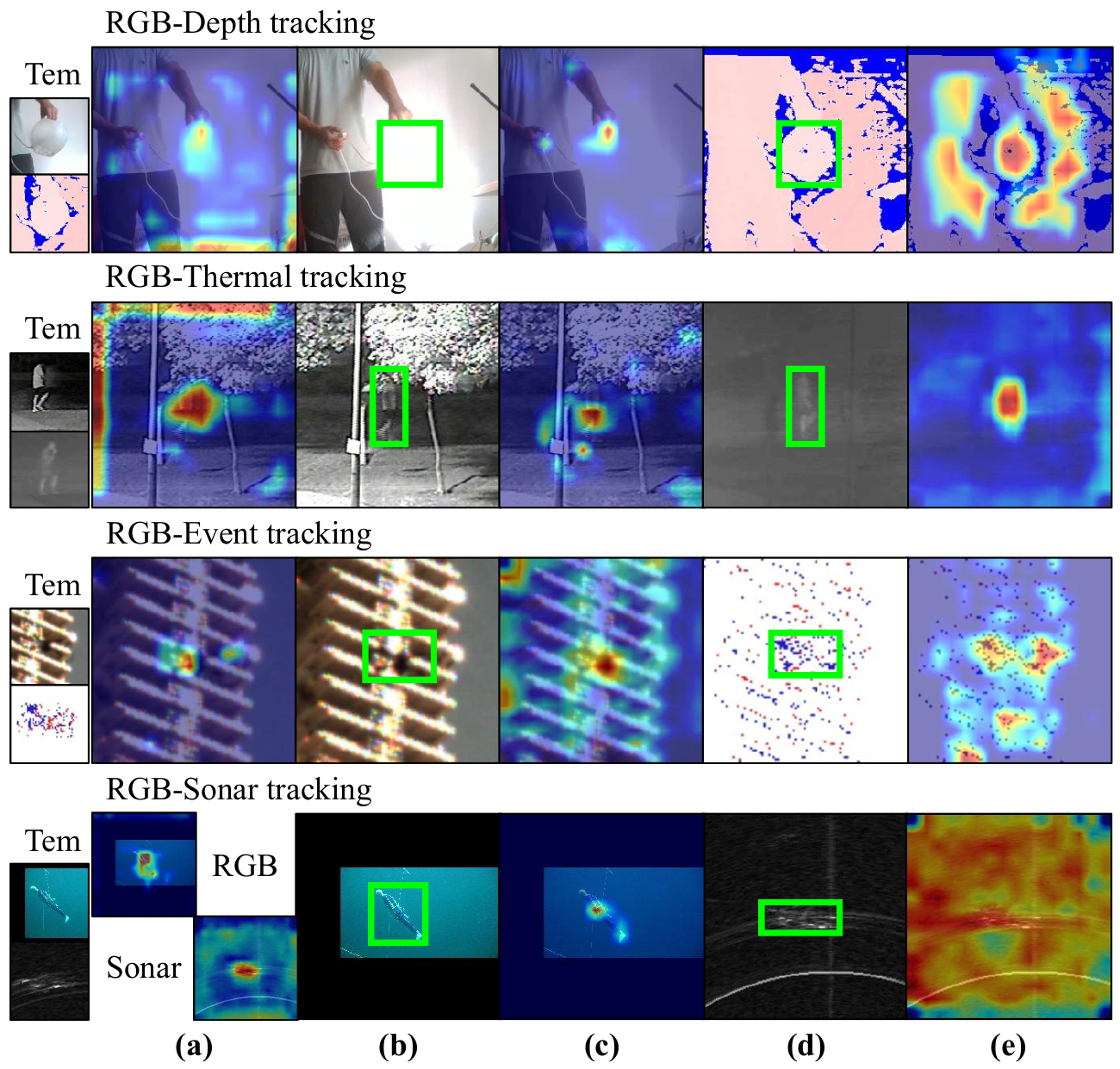}
\caption{Visualization of LightFC-X heat maps. (a) Heat maps of cross-modal feature. (b) RGB modality search areas. (c) Heat maps of RGB feature. (d) X modality search areas. (e) Heat maps of X modality feature.}
\label{fig: heatmap}
\end{figure}

 To explore how our model works, we visualize the heat maps before and after cross-modal interaction in Fig.~\ref{fig: heatmap}. Our method effectively improves feature discrimination in challenging multimodal tracking scenarios.

\section{Conclusion}
\label{sec:conclusion}
In this work, we present a lightweight RGB-X (Thermal, Depth, Event, and Sonar) tracker named LightFC-X, which achieves an optimal balance of parameters, performance, and speed. It includes two novel modules: an efficient cross-attention module (ECAM) and a spatio-temporal template aggregation module (STAM). The ECAM module achieves efficient spatial integration of cross-modal features. The STAM module achieves spatiotemporal aggregation of target appearance templates through module fine-tuning paradigm. Comprehensive experiments show that the proposed LightFC-X achieves state-of-the-art performance among lightweight trackers across several multimodal tracking benchmarks. We expect this work can attract more attention to lightweight multimodal tracking.

\textbf{Limitations.} Although the LightFC-X family achieves a unified lightweight framework, its use of spatiotemporal features remains deficient. With the emerging of sequence-training paradigms, we are interested in how to use the paradigm to enhance the use of spatiotemporal information by lightweight multimodal trackers.



{\small
\bibliographystyle{ieeenat_fullname}
\bibliography{main}
}


\end{document}